\documentclass[runningheads]{llncs}

\usepackage{eccv}

\usepackage[final]{graphicx}
\usepackage{microtype}
\usepackage{forest}
\usetikzlibrary{%
shadows,%
} 
\usepackage{xcolor}
\usepackage{relsize}
\usepackage{booktabs}
\usepackage{paralist}

\usepackage[breaklinks,colorlinks,citecolor=black,linkcolor=black]{hyperref}
\usepackage{orcidlink}

\usepackage{commenting}
\usepackage{csquotes}
\usepackage{fontawesome}
\setkeys{Gin}{draft=false} %

\let\highlight\textbf

\newcommand*{\Concept}[1]{\textsf{#1}}

\newcommand{\narrowbold}[1]{%
  {\fontseries{b}\selectfont#1}%
}

\newcommand*{\Challenge}[1]{{%
    \textsuperscript{\color{YellowOrange}\faTrophy}\narrowbold{#1}}}
\newcommand*{\Proposal}[1]{{%
    \textsuperscript{\color{SeaGreen}\faLightbulbO}\emph{#1}}} 

\title{Concept-Based Explanations in Computer Vision: Where Are We and Where Could We Go?}

\titlerunning{Concept-Based Explanations in CV: Where Are We \& Where Could We Go?}

\author{
Jae Hee Lee\inst{1}\textsuperscript{*}\orcidlink{0000-0001-9840-780X}
\and Georgii Mikriukov\inst{2}\orcidlink{0000-0002-2494-6285}
\and Gesina Schwalbe\inst{3}\orcidlink{0000-0003-2690-2478}
\and Stefan Wermter\inst{1}\orcidlink{0000-0003-1343-4775}
\and Diedrich Wolter\inst{3}\orcidlink{0000-0001-9185-0147}
}

\authorrunning{J.~H.~Lee et al.}

\institute{
University of Hamburg, Germany\\
\email{\{jae.hee.lee, stefan.wermter\}@uni-hamburg.de}
\and
Anhalt University of Applied Sciences, Germany\\
\email{georgii.mikriukov@student.hs-anhalt.de}
\and
University of Lübeck, Germany\\
\email{\{gesina.schwalbe, diedrich.wolter\}@uni-luebeck.de}
}

\begin{document}

\maketitle
\def\thefootnote{*}\footnotetext{{Author names are in alphabetic order.}}
\def\thefootnote{\arabic{footnote}}

\begin{abstract}
    Concept-based XAI (C-XAI) approaches to explaining neural vision models are a promising field of research, since explanations that refer to concepts (i.e., semantically meaningful parts in an image) are intuitive to understand and go beyond saliency-based techniques that only reveal relevant regions. Given the remarkable progress in this field in recent years, it is time for the community to take a critical look at the advances and trends. Consequently, this paper reviews C-XAI methods to identify interesting and underexplored areas and proposes future research directions. To this end, we consider three main directions: the choice of concepts to explain, the choice of concept representation, and how we can control concepts.
    For the latter, we propose techniques and draw inspiration from the field of knowledge representation and learning, showing how this could enrich future C-XAI research.

    \keywords{Concept-Based Explainable AI \and Concept Embedding Analysis \and Concept Control \and Neuro-Symbolic AI \and Knowledge Representation}
\end{abstract}

\section{Introduction}

As the capabilities of deep learning models grow and as our society uses them more, it becomes increasingly important to \emph{understand} how they work~\cite{saeed2023explainable} and how to \emph{control} them in effective ways \cite{garcez2023neurosymbolic}: Understanding how a model works is the basis for trusting the model~\cite{kaur_trustworthy_2022} and for its verification against ethical, privacy, or safety requirements~\cite{euaiact}. Control is imperative to effectively enforce requirements by design, during maintenance, or through manual ad-hoc intervention. Understanding and control have formed the basis for a wealth of explainable artificial intelligence (XAI) methods for computer vision (CV) \cite{zhangVisualInterpretabilityDeep2018,das_opportunities_2020,schwalbe2023comprehensive}.

In XAI for CV, early post-hoc explainability approaches have focused on areas of importance of features in the input image relevant for a vision model's decision \cite{ancona_gradientbased_2019,liang2021explaining}. However, these approaches do not explain what happens internally in the model. Concept-based XAI (C-XAI) \cite{schwalbe_concept_2022,lee_neural_2024,poeta_concept-based_2023} overcomes this shortcoming by explaining how a vision model represents input in its intermediate layers using semantically meaningful concepts that can be understood by users. Finding concept-based descriptions of internal representations is needed to gain more insight into the internal information processing of the model \cite{rauker2023transparent}, since concepts can act as a Rosetta Stone, i.e., as a common alphabet between users and the model. Such concepts can be task-related objects (e.g., \Concept{head}, \Concept{beak}) or scene properties (e.g., \Concept{red}, \Concept{sunny}), and are not necessarily part of the output labels.

\begin{figure}[t]
    \centering
    \includegraphics[width=.9\textwidth]{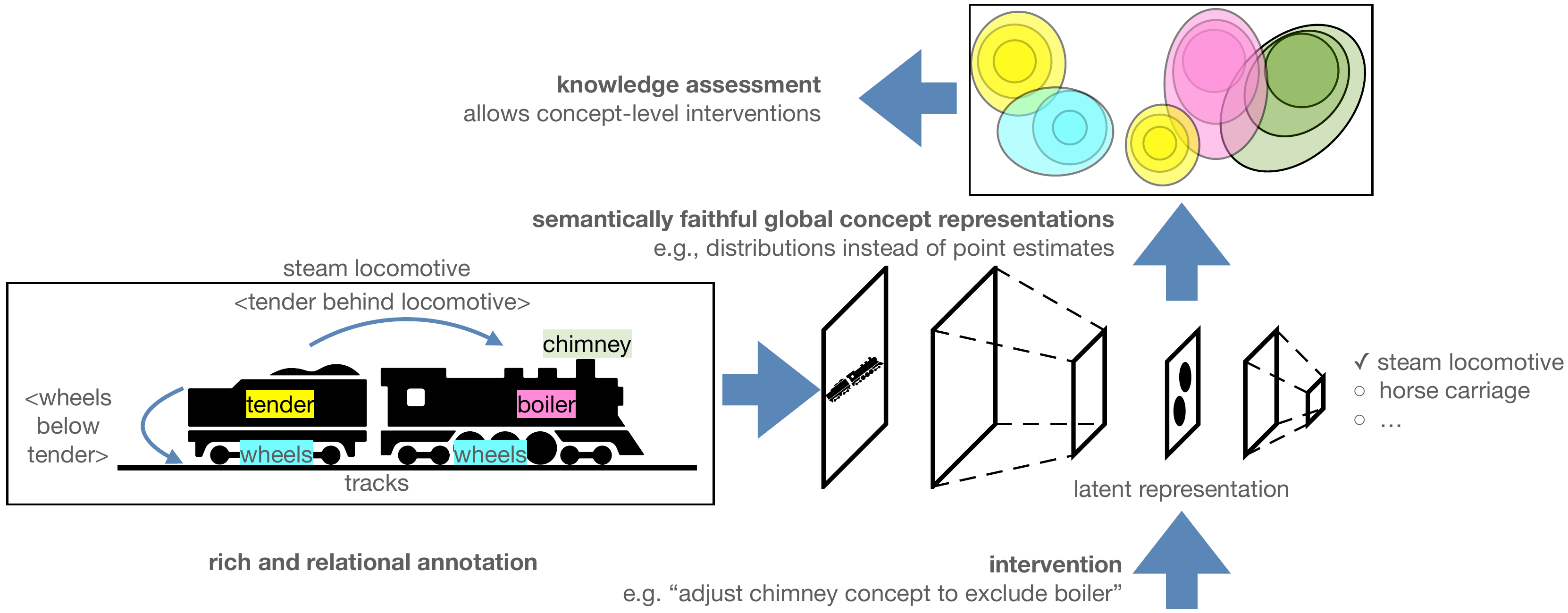}
    \caption{\label{fig:overview} Overview of envisaged methodology for model understanding and control.
      Using rich and relational concept annotations (e.g., grounded in an ontology) of visual model inputs, intuitive concepts and relations are associated with global, expressive, and semantically faithful concept representations in the model's latent space (e.g., distributions). This allows interactive knowledge verification and local or global control, e.g., adjusting the concept representation to globally separate the concept \Concept{boiler} from the concept \Concept{chimney}. }
\end{figure}

\paragraph{\textbf{Goal and contributions.}}
Our overall aim is to give XAI researchers a good starting point to dive into the subtopic of C-XAI and a guide to interesting next steps to advance the field further. Previous C-XAI surveys \cite{schwalbe_concept_2022,poeta_concept-based_2023,lee_neural_2024,rauker2023transparent} have focused mainly on motivating and introducing C-XAI methods, focusing less on discussing the future of C-XAI research. In this paper, in addition to \textbf{reviewing the state of the art in C-XAI}, our objective is to draw attention to important challenges of C-XAI, which are largely neglected in the literature. We \textbf{discuss the state of the art and open challenges} in extracting new \textbf{concept types}, devising \textbf{concept representations} that go beyond the initial vector-based approach, and applying \textbf{concept control} mechanisms, where each \Challenge{challenge} (a trophy icon followed by bold text) is accompanied by \Proposal{proposals} (a light bulb icon followed by italicized text) on how to tackle it.

Our particular position in this paper is that C-XAI can benefit from the established field of knowledge representation and reasoning (KR) \cite{vanharmelen2007handbook,brachman_knowledge_2014}. This includes verifying and controlling the \textbf{ontological commitment}~\cite{bricker_ontological_2016} of a vision model, that is, whether it has learned the right concepts and their relations to each other. Answering these questions empowers the pipeline in \autoref{fig:overview}.

\paragraph{\textbf{Scope.}}
We will only briefly touch on the mature C-XAI directions for investigating concept relevance scores \cite{kim2018interpretability,goyal2019explaining,graziani2020concept,yeh2020completenessaware,pfau_robust_2021}, and leave aside the promising applications of combination with feature importance methods \cite{motzkus2024locally,achtibat2023attribution}, and guidance for creating counterfactual explanations \cite{motzkus2024coladce}. Similarly, with regard to challenges, we do not discuss in detail the already well-known issues of concept completeness \cite{yeh2020completenessaware,sawada2022concept,chen2020concept}, concept leakage \cite{mahinpei2021promises,kazhdan2021disentanglement,marconato2023interpretability,hoffmann2021this,marconato2022glancenets,havasi2024addressing}, lack of causality of concepts (\Concept{fluffy} \& \Concept{ear} not necessarily implies the composite \Concept{fluffy ear}) \cite{lovering2022unit}, as well as cost and availability of concept labels \cite{belem2021weakly,oikarinen_label-free_2023}. Instead, we identify and highlight the so far underexplored directions for potential advancements. Also, there is the question of how to evaluate the quality of C-XAI methods \gsadd{\cite{espinosazarlenga2022concept}}, which goes along with the general issues in XAI to define meaningful functionally-, human- and application-grounded metrics \cite{schwalbe2023comprehensive,li2020quantitative,doshi-velez2017rigorous} and is not considered here.



In the next section, we will give a compact review of C-XAI and relevant background (\autoref{sec:conc-based-expl}), and then present our results on open challenges along the dimensions of concept type (\autoref{sec:concept-extraction}), representation (\autoref{sec:conc-repr}), and control (\autoref{sec:concept-control}). 

\section{Background on C-XAI for Computer Vision}
\label{sec:conc-based-expl}

Concept-based explainable AI seeks to enhance the interpretability of AI models by connecting their internal representations with human-under\-stand\-able \emph{concepts}. The definition of the concept used here varies across the literature.

\subsection{What is a concept?}

Poeta et al.~\cite{poeta_concept-based_2023} define concepts as ``human-interpretable high-level features of the input data that are important for the model’s decision-making process''. This definition highlights the connection to features as used in feature importance methods.
In semantic contexts, a concept is generally a notion that can be described using natural language \cite{schwalbe_concept_2022}, for example, a synonym set in the lexical database WordNet \cite{fellbaum1998wordnet} (e.g., \Concept{ear}, \Concept{fluffy}) or a combination thereof (e.g., \Concept{fluffy ear}). This definition focuses on close alignment with human language and is the most commonly used definition in C-XAI~\cite{fong2018net2vec,kim2018interpretability,bau2017network}. We use this as the default notion in this paper. Extending this to image analysis, a concept can be considered as a meaningful region within an image \cite{fong2018net2vec,ghorbani2019towards,schwalbe_concept_2022}. 
So far, concepts considered are only vaguely interrelated and do not capture the rich structure of an ontological language that allows to define complex concepts from a set of basic ones.

In symbolic AI literature, particularly within KR, concepts are viewed as what can modeled as logic predicates, characterized by their relations to other concepts \cite{baader:09,guarino1998formala}. This structural approach emphasizes the logical relationships and hierarchy among concepts but is underexplored in C-XAI \cite{schwalbe2022enabling}. 

\subsection{An Overview of C-XAI Directions for CV}\label{sec:c-xai-survey}

In this section, we briefly review the state-of-the-art of C-XAI methods, to set the scene for later discussion (see \autoref{fig:taxonomy} in Appendix \ref{sec:appendix-taxonomy} for a taxonomy).
For further details, the reader is referred to more elaborate surveys \cite{schwalbe_concept_2022,lee_neural_2024,poeta_concept-based_2023,rauker2023transparent}.

C-XAI aims to associate the mentioned human-understandable concepts with a representation allocated in a neural model's latent space(s), i.e., the intermediate output space of the model. We use the terms \emph{concept representation}, or equivalently, \emph{concept embeddings}, as a collective notion to encompass the variety of existing methods to represent concepts. In order to be understandable, a model must operate using the same conceptual ``alphabet'' as humans. Ideally, black-box models should also internally represent and use concepts that match those from the catalog of human cognition.

Concept representations can be \emph{post-hoc} extracted (i.e., after training a model) or \emph{ante-hoc} enforced (i.e., explainable by design) \cite{schwalbe_concept_2022,poeta_concept-based_2023,lee_neural_2024,rauker2023transparent }. We can also categorize C-XAI into \emph{supervised} and \emph{unsupervised} methods \cite{schwalbe_concept_2022}: Supervised methods utilize pre-defined concept specifications, such as labeled concept examples, to check whether a neural model encodes information about a concept in question. Unsupervised methods instead aim to identify what concepts a model has learned; considerations here are what qualifies a representation as that of a concept, typically cluster centers \cite{ghorbani2019towards,zhang2021invertible} or linear basis directions \cite{zhang2021invertible} (cf.~\autoref{fig:concept-repr-variants}); and how to ensure human interpretability of the found concepts, e.g., by constraining found concepts to be connected image regions \cite{ghorbani2019towards}. In the following, we review existing variants for concept representation and discuss supervised and unsupervised C-XAI methods in detail.

\subsubsection{Concept Representation Variants.}
\begin{figure}[t]
    \includegraphics[width=0.42\linewidth]{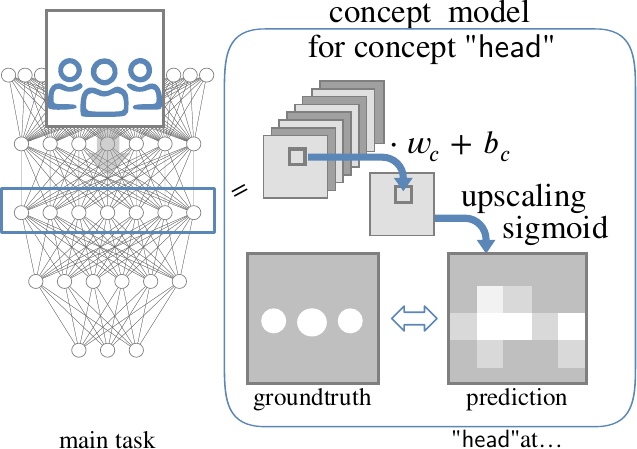}\hfill
    \includegraphics[width=0.28\linewidth]{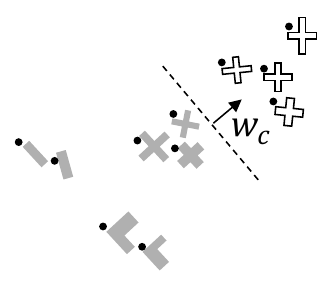}%
    \includegraphics[width=0.28\linewidth]{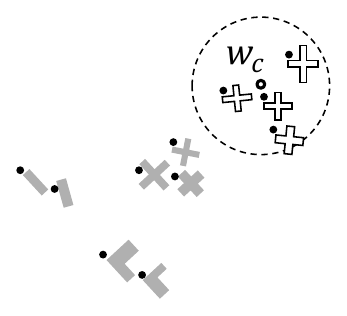}%
    \caption{Illustration of Net2Vec~\cite{fong2018net2vec} for associating a concept with a linear separator with weight vector $w_c$ in (activation pixel) latent space (\emph{left}), and illustration of typical concept representation variants (\emph{center:} direction-based, \emph{right:} cluster-based).
    }
    \label{fig:concept-repr-variants}
\end{figure}

A concept representation consists of two parts: the representation of the human-interpretable part (usually via examples \cite{bau2018gan}, in vision-language models also via text \cite{parekh2024conceptbased,liang2022multiviz}) and the associated latent representation, which is typically given by the parameters of the function that associates a concept with its latent representation. For example, TCAV~\cite{kim2018interpretability} defines a concept via images with binary classification labels, and the association function as a binary classifier of latent vectors.
In its simplest form, a concept is associated with a single unit of the network (neuron~\cite{koh2020concept} or filter~\cite{bau2017network} in a given layer). A more general perspective represents concepts by weight vectors with one weight per network unit of interest, taking the role of directions or centroids in latent space (see \autoref{fig:concept-repr-variants}).
Such techniques were shown to capture better the distributed way \cite{craven1992visualizing} of how information is stored in a model and were established in TCAV~\cite{kim2018interpretability} and Net2Vec~\cite{fong2018net2vec}. Since then, more complex representations include clusters~\cite{ghorbani2019towards,posada-moreno2024eclad}, and kernel functions~\cite{crabbe2022concept}. It should be noted that nonlinear association functions are also investigated, such as generalized linear models \cite{alvarez-melis2018robust,marcinkevics2020interpretable}, or normalizing flows \cite{esser2020disentangling,rombach2020making}. However, this sacrifices the interpretability of the association \cite{kim2018interpretability}.
The selection of the association function is task-specific, focusing on aspects such as the concept type \cite{schwalbe_concept_2022} like spatial localization (e.g., image classification~\cite{kim2018interpretability}, segmentation~\cite{fong2018net2vec}), and concept values (e.g., binary \cite{kim2018interpretability}, regression~\cite{graziani2020concept}, multi-class \cite{kazhdan2020now}); and constraints on the involved latent representations like being non-negative \cite{zhang2021invertible},
unit vectors \cite{bau2017network} or even a complete orthogonal basis \cite{koh2020concept,chen2020concept,yuksekgonul2022posthoc}.

Following an initial \cite{kim2018interpretability} and still prevalent \cite{goyal2019explaining,graziani2020concept,yeh2020completenessaware,pfau_robust_2021} application of C-XAI, some authors also demand as part of the concept representation an importance score \cite{posada-moreno2024eclad}. This score tells how much the concept participates in the model's decision process \cite{posada-moreno2023scalable,posada-moreno2024eclad}, which is similar to feature importance~\cite{baehrens2010how}.

\subsubsection{Supervised Concept Analysis.}

Supervised concept embedding analysis methods associate predefined concepts with the units of the neural model. 
First approaches matched concept segmentations to the most similarly activated convolutional neural network (CNN) filters \cite{bau2017network}. Fong et al.\ in Net2Vec~\cite{fong2018net2vec} and Kim et al.\ in TCAV~\cite{kim2018interpretability} soon after trained linear models, for concept segmentation and concept classification respectively, to separate concept 
from non-concept activations, with their weight vector serving as concept embedding vector. This is up to now the basis for essentially all post-hoc supervised techniques: Their linear models were extended to linear regression \cite{graziani2018regression,graziani2020concept}, kernel-based methods producing region-based concepts \cite{crabbe2022concept}, and from global to image-local explanations by training on concept data subsets \cite{mikriukov2023gcpv,zhang2018examining}.

By contrast, ante-hoc (or explainable-by-design) approaches typically use the simple representation again and associate single units in a layer with concepts. They were first introduced as concept bottleneck models (CBMs) \cite{koh2020concept,losch2019interpretability}. This was later improved by denoising techniques to model concept interdependencies \cite{bahadori2020debiasing,havasi2024addressing}, semisupervised training strategies for label efficiency \cite{belem2021weakly,oikarinen_label-free_2023}, concept hierarchies \cite{marcos2020contextual}, binary \cite{havasi2024addressing} and multidimensional \cite{espinosazarlenga2022concept} concept representations; and combined with unsupervised methods \cite{sawada2022concept} to overcome the well-known challenge of choosing a complete set of concepts, that is, one sufficient for the task \cite{yeh2020completenessaware,sawada2022concept,chen2020concept}. 
Furthermore,
CBMs are criticized for concept-leakage \cite{mahinpei2021promises,kazhdan2021disentanglement,marconato2023interpretability,hoffmann2021this,marconato2022glancenets}: The vector produced by all concept neurons may learn to encode not only information about the given concepts but also \enquote{leak} other information to achieve higher accuracy.

\subsubsection{Unsupervised Concept Analysis.}


Unsupervised concept analysis methods identify the most important concepts in the feature space without labeling information. 
They achieve concept completeness by design, but at the cost of possibly uninterpretable concept formations. 
The manual labeling effort for assigning labels to the found concepts is still necessary to finally establish the concept association.
Techniques to identify prevalent features include standard clustering of activations obtained from a probing dataset, as first done for image-level concepts \cite{ghorbani2019towards,ge2021peek}; and via (multi-layer) activation pixel clustering for image-region concepts \cite{posada-moreno2024eclad,posada-moreno2023scalable}. This was shown to be subsumed by matrix factorization techniques such as k-means clustering, classical PCA, or non-negative matrix factorization~\cite{vielhaben2023multidimensional,zhang2021invertible,fel_holistic_2023,leemann2023when,fel2023craft},


There also exist ante-hoc methods that, similar to CBMs, have a bottleneck layer. Instead of assigning one neuron per concept, they learn to encode concepts as prototype vectors. Comparing these with the intermediate representations produces the concept scores. This case-based reasoning approach was first introduced in ProtoPNet \cite{chen2019this,li2018deep}  and continued in its successors on object detection \cite{feifel2021reevaluating}, with prototype sharing across output classes \cite{rymarczyk2021protopshare}, and with preferable cosine similarity instead of $L_2$ distance for prototype comparison \cite{willard2024this}. 

\subsection{Ontological Commitment in Knowledge Representations}
We will here show how the notion of ontological commitment from the field of knowledge representations naturally translates to C-XAI requirements, which are further elaborated later.
For everyday concepts, humans typically have an understanding of a concept based on \emph{what other concepts are related and 
via which relations}. 
To connect this to neural models, note that also the model's internals are supposed to be a (learned) knowledge representation. Thus, both concepts and their relations induce \emph{constraints} on valid model (intermediate) predictions \cite{schwalbe2022enabling}. For example, consider object existence constraints from object-to-part relations \cite{giunchiglia2022roadr,schwalbe2022enabling}: since the \Concept{head} is part of a \Concept{person}, the presence of a \Concept{head} should imply the presence of a \Concept{person}. Similarly for hierarchical class subsumption \cite{roychowdhury2018image}: since a \Concept{human} is a \Concept{movable object}, the detection of a \Concept{person} implies it may be \Concept{movable}.
Aside from these constraints, we also expect explanations to be more intuitive for humans to understand if they use humans' cognitive catalog of concepts and relations (also known as cognitive chunks \cite{doshi-velez2017rigorous}). Therefore, an implicit requirement for concept representations is that they support reasoning with concepts and capture human prior knowledge about the task. 
The respective kind of reasoning is determined by the so-called \emph{ontological commitment} \cite{davis_whatisKR:93}. 
Ontological commitment refers to the catalog of defined \emph{concepts} (1-ary logic predicates) as well as \emph{relations} (binary, possibly n-ary predicates), for example, \texttt{IsSimilarTo}, \texttt{IsSubclassOf}, \texttt{IsPartOf}, \texttt{IsCloseTo}. The term \emph{commitment} signals the choice of admissible concepts, and 
significantly influences what kinds of inferences are possible or easy. For example, if \Concept{dog} and \Concept{cat} are organized as subconcepts of concept \Concept{pet}, then their co-occurrence with humans is easier to predict than choosing a zoology-motivated taxonomy (cf.~\autoref{fig:class-hierarchies}).

\section{Types of Concepts}
\label{sec:concept-extraction}



At the heart of the problem definition in C-XAI lies the questions of what concepts to extract and where to extract them from. In the visual domain, multiple concept types have already been considered \cite{schwalbe_concept_2022}: image-level scene attributes (e.g., \Concept{sunny}) \cite{bau2017network} and image qualities (e.g., \Concept{contrast}) \cite{abid2022meaningfully}; as well as attributes of image regions such as object (e.g., \Concept{person}) and object part classes (e.g., \Concept{beak}) \cite{bau2017network,koh2020concept}, and object attributes such as material, texture \cite{kim2018interpretability}, and color \cite{schauerte2010google512}.

Apart from a few exceptions \cite{marcos2020contextual}, the concepts are based on layered neural networks or spatial alignment in unimodal CNNs. Hence, post-hoc C-XAI has in CV so far been applied to classifiers \cite{kim2018interpretability,fong2018net2vec}, regression \cite{graziani2018regression}, object detectors \cite{schwalbe2021verification,mikriukov2023evaluating},
and only recently for the first time to video models \cite{saha2024exploring,ji2023spatialtemporal};
applications to language models \cite{zhao2024explainability}, generative adversarial networks \cite{bau2018gan}, and quite recently to diffusion models \cite{gandikota_unified_2024,ismail2023concept} suggest that more architectures could be covered. An example is the Vision Transformer (ViT), which has recently become a popular CV architecture. Its self-attention mechanism, however, is difficult to interpret.
Rigotti et al.~\cite{rigotti2021attention} propose the Concept-Transformer, which extends attention from low-level features to high-level concepts, providing plausible and faithful explanations. 

\subsection*{Open Challenges}
\label{sec:ct-open-challe-nges}

C-XAI Research so far seems to be limited to the mentioned static attributes of images and image regions that are extracted from CNNs. This neglects concepts arising from \emph{temporal} or \emph{other sensory features}, as well as \emph{other architectures} such as ViTs \cite{dosovitskiy2020image}. 
Since they could take important roles in future critical applications, we argue that more research is needed on concept extraction in these fields. 

\subsubsection{Temporal and Multimodal Concepts.}

What remains largely unexplored is \Challenge{the identification of concepts for temporal and other sensory features}, despite being of interest for many important robotic applications like automated driving. These have an inherent temporal and multisensory resolution, which is inevitable for reliable prediction of trajectories, e.g., to differentiate advertisements on trucks from true pedestrians. Meanwhile, research of C-XAI in videos is very sparse, with the first TCAV-based work still concentrating on objects instead of movement patterns as concepts \cite{saha2024exploring,ji2023spatialtemporal}. Similarly, the investigation of C-XAI in multimodal models has just started, but with a focus on vision-language models to utilize the language input for concept definition \cite{parekh2024conceptbased,liang2022multiviz}. \Proposal{Since it is possible to disentangle multimodal representations into single-modal ones \cite{lyu2022dime}, this might be an attack point for the transferral of C-XAI techniques to multimodal non-language models.} Generally, it would be interesting to see how and what concepts are represented in video and/or multimodal models, in order to enable in-depth debugging.
Furthermore, \Challenge{the so-far unused temporal resolution in spatio-temporal concepts might open up new ways of self-supervised concept extraction.} It is well known that motion cues such as optical flow arising from temporal consistency in real-world videos are valuable information for object segmentation \cite{Yang_2021_ICCV,Xiong_2021_ICCV}. This has, to our knowledge, not yet been used to analyze trained latent representations of video processing models. \Proposal{Latent representations that occur in spatio-temporal regions with stable optical flow, such as on a moving object, might be interpreted as learned object properties.} That would, for example, allow self-supervised part-object extraction, to validate whether force exertion (e.g., \Concept{locomotive} tugging \Concept{tender} \autoref{fig:overview}, swarm-like behavior), connectedness (e.g., \Concept{arms} typically do not detach), or even shadows on 3D objects are adequately modeled.

\subsubsection{Concepts in New Architectures.}

As reviewed above, it is not yet clear \Challenge{how to associate concepts in new architectures.} ViTs, for example, break with the direct association of neurons with spatial locations in the input. This, however, is utilized in nearly all C-XAI methods for extraction of subimage concepts: A concept in a spatial position must be reflected in the activation spatially aligned to that position, as already done in the base C-XAI methods \cite{fong2018net2vec, zhang2021invertible, losch2019interpretability, feifel2021reevaluating}. An alternative would be to use full-layer concept vectors, and during inference allocate them to individual image input regions by feature importance techniques, as done in \cite{lucieri2020explaining} but with mediocre success. \Proposal{A combination that leverages the coarse patch-wise processing of vision transformers together with feature attribution methods may be a promising direction.} 
Stassin et al.~\cite{stassin2023explainability} discuss adapting existing XAI techniques to Transformers by converting embeddings into pseudo-activation maps, with a particular interest in applying this approach to the MLP layers
Another approach is \Proposal{training a sparse autoencoder} on the activations of a layer, which is so far used in understanding large language models~\cite{huben_sparse_2024} and could be transferred to the vision domain.
Similarly to ViTs, explaining diffusion models for image generation has only just sparked interest, both ante-hoc \cite{ismail2023concept} as well as post-hoc \cite{park2024explaining,gandikota_unified_2024}, although diffusion models are already being used in turn for concept discovery \cite{varshney2024generating}. That could be an entry point for the diffusion model analysis.
In summary, we see many opportunities to advance our understanding of novel model types via C-XAI.

\section{Concept Representation}
\label{sec:conc-repr}
Concept representation encompasses two directions: How a specific concept is represented and which concepts are represented.

\begin{figure}[tb]
    \begin{subfigure}{.38\linewidth}
        \centering
        \includegraphics[width=.95\linewidth] {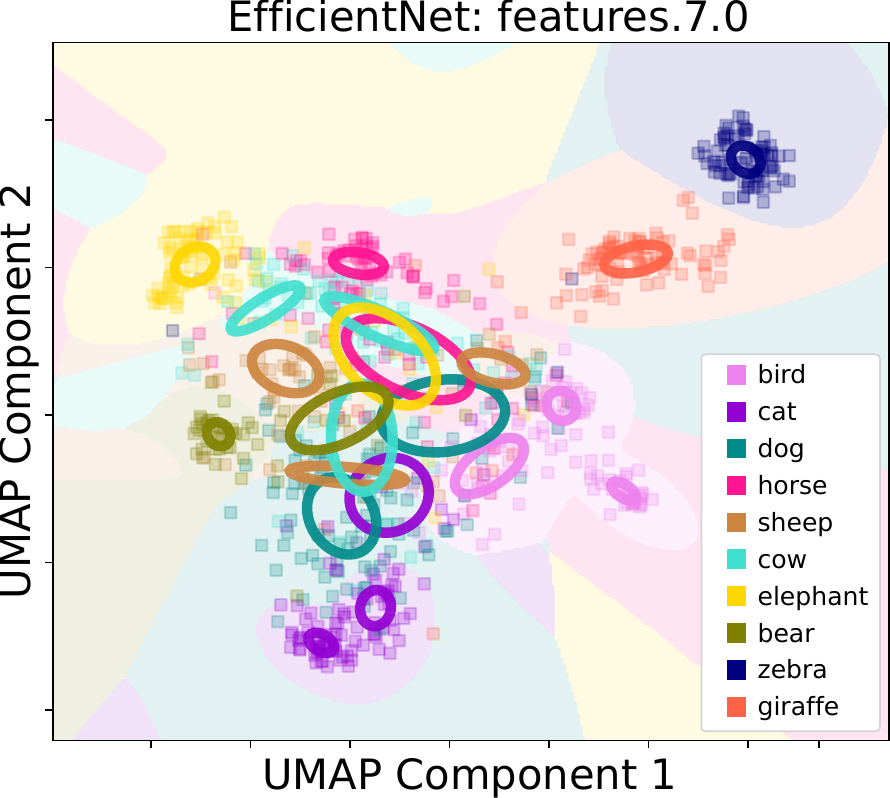}
        \caption{
        Complex distribution of image-local concept representations in EfficientNet-B0's~\cite{tan2019efficientnet} last layer
        \emph{Ellipses} \& \emph{shades} indicate fitted Gaussians. Details: Appendix~\ref{sec:appendix-graphics}.
      }
      \label{fig:gcpv-distribution}
  \end{subfigure}%
    \hfill%
    \begin{subfigure}{.60\linewidth}
        \centering
        \includegraphics[width=\linewidth]{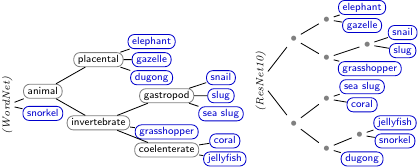}%
        \caption{Comparing class hierarchies defining the subsumption relation, as presented in \cite[Fig.~4]{wan2020nbdt}. \emph{Left:} extracted from lexical database \highlight{WordNet}~\cite{fellbaum1998wordnet}, \highlight{zoology-motivated}; \emph{right:} extracted from \highlight{ResNet10}'s last layer~\cite{he2016deep} by hierarchical clustering of concept embedding vectors \cite{wan2020nbdt,mikriukov2023gcpv}, motivated by \highlight{visual similarity} of typical backgrounds.}
        \label{fig:class-hierarchies}
    \end{subfigure}%
  \caption{Illustration of the ontological commitment (\autoref{fig:class-hierarchies}, \emph{right}), and complex concept distribution (\autoref{fig:gcpv-distribution}, \emph{left}) in actual vision model's latent spaces.}
\end{figure}


\subsection{Basic Types of Concept Representations.}

Using single neurons (i.e., unit vectors in latent space) as concepts~\cite{koh2020concept} makes it easy to quantify concept attribution via neurons' activation magnitude. However, this can be overly simplistic~\cite{margeloiu2021concept} because it overlooks the distributed nature of neural network representations~\cite{kim2018interpretability,fong2018net2vec,craven1992visualizing}.
%
Standard now are vector-based concept representations that hold weights for each neuron \cite{kim2018interpretability,ghorbani2019towards} or filter \cite{fong2018net2vec,mikriukov2023gcpv,zhang2021invertible} of one or several \cite{posada-moreno2024eclad} layers. They require optimization but provide more accurate concept embeddings \cite{fong2018net2vec,kim2018interpretability}. So far, only a few exceptions generalize this from global point estimates to (nonlinear) subspaces \cite{esser2020disentangling}, latent space regions \cite{crabbe2022concept}, or hierarchies of (local or global) point estimates \cite{mikriukov2023gcpv,wan2020nbdt,marcos2020contextual}. In the following subsection, we will argue the immediate shortcomings of the currently prevalent vector-based representations.

\subsection{Ontological Commitment of Concept Representations.}

The available background commonsense knowledge regarding concept definitions (e.g., \texttt{IsPartOf(\Concept{head},\Concept{person})}) is essential for pinning down semantics. 
Manually crafted, large ontologies often aim to capture the ontological commitment of human common sense. Notable examples are WordNet~\cite{fellbaum1998wordnet}, Cyc~\cite{lenat1989building}, SUMO~\cite{niles2001standard}, or ConceptNet~\cite{speer2017conceptnet}. 
To connect these sources of information to C-XAI one has to ground ontology concepts in network activation. 
As a first step,  individual concepts are grounded in network activation, but it is desirable to extend this approach to capture more expressive ontological languages, e.g., \cite{oezcep-alc-JAIR:23}. 
With respect to grounding individual concepts, 
recall that, e.g., in 
TCAV~\cite{kim2018interpretability} and Net2Vec~\cite{fong2018net2vec} the cosine similarity was used as a measurement for semantic similarity of latent concept representations, and vector addition as semantic combination of concepts (a kind of logical \texttt{AND}). This can now be considered as relations in the ontological language of their chosen vector-based concept representation. Probing what concept representation is a combination of others (e.g., $\Concept{wood}+\Concept{green}\approx\Concept{tree}$\cite{fong2018net2vec}) or is similar to others (e.g., $\Concept{brown hair}\sim\Concept{black hair}$\cite{kim2018interpretability}) extracts the constraints and hence the ontological commitment of what the model has learned. This commitment, however, does not necessarily coincide with human intuition but can encode unwanted biases like $\Concept{apron}\sim\Concept{female}$. 
Therefore, the important goal of C-XAI to uncover faulty learned knowledge reformulates as to \emph{verify, validate, and control the ontological commitment and conceptualization of vision models}. 

Unfortunately, little investigation has been devoted so far to the ontological commitment of types of concept representations. Vanilla (point-estimate) vector embeddings allow measurement of semantic similarity via cosine distance \cite{fong2018net2vec} but are criticized for their inability to model richer concept relations \cite{gutierrez-basulto18from,xiong_faithful_2022}.
Spatial calculi \cite{schockaert2008fuzzy} become applicable to concept segmentations for modeling object-part-relations on image regions \cite{schwalbe2022enabling} or region-based concept representations (instead of point estimates) to model subsumption relations \cite{mikriukov2023gcpv,wan2020nbdt} (extractable via hierarchical clustering, cf.~Figs.~\ref{fig:class-hierarchies}, \ref{fig:gcpv-distribution}).
Donadello et al.~\cite{donadello2017logic} showed that neural networks are also capable of learning more complex relations.
This is in accordance with the findings that deep neural networks employ simple reasoning steps on concepts across several layers, the subnetworks encoding these also called circuits \cite{olah2020zoom}.

\subsection*{Open Challenges}
We will now first argue, why the prevalent vector-based representations fall short of capturing some basic interesting information about concepts.
This is then extended to the perspective of ontological commitment, where proposals are made to find richer relations between concept representations for better model validation and verification.

\subsubsection{Questioning Point Estimates as Concept Representations.}


A challenge posed by the prevalent vector-based approaches is their two inherent assumptions that we will question in the following: (i) Concepts can meaningfully be approximated by \emph{linear} trajectories in latent space pointing from \Concept{less concept} to \Concept{more concept} \cite{kim2018interpretability,pahde2024navigating}, and (ii) this direction can be expressed by a point estimate.

\Challenge{Linear point estimated representations are too simplistic, concepts should be modeled by regions or distributions.} This can be argued from two perspectives. For one, while point estimates might be sufficient for small models and datasets with few clearly distinct concepts \cite{koh2020concept}, Mikriukov et al.~\cite{mikriukov2023evaluating} showed that this can break down at scale, as illustrated in \autoref{fig:gcpv-distribution}: Concepts in larger object detectors are smeared over the latent space at different densities, start overlapping, and even break down into distinct subconcepts. Such relevant information cannot be captured by point estimates. Instead, region-based \cite{crabbe2022concept,posada-moreno2023scalable,ozcep_embedding_2023} or density-based \cite{mikriukov2023gcpv} approaches can capture spread (or even density and thus outliers), non-connectedness (i.e., sub-concepts), and overlap (i.e., concept confusion or concept commonalities) of concepts in the model's latent spaces. \Proposal{Future research could involve generalizing local C-XAI approaches \cite{mikriukov2023gcpv,zhang2018examining}, fitting Gaussian mixture models to sets of such local concept vectors} and investigating factors that influence concept spread. Furthermore, \Proposal{the region-based approach poses an interesting direction}. For example, representing concepts as \emph{cones} could be promising, as they naturally come with negation, intersection (\texttt{AND}) and union (\texttt{OR}) on concepts \cite{oezcep-alc-JAIR:23,leemhuis-oezcep-JAR:23}, as picked up again below.

%
As a second argument, we would like to draw attention to \Challenge{modeling the rate of change}: So far, C-XAI-based approaches only considered the general direction towards \Concept{more concept} regions but not the rate of change when traversing the trajectory. It might, however, be interesting information whether the model assumes a rapid change (turning point) like one would expect for \Concept{glasses} versus \Concept{broken glasses}; a somewhat smooth transition, like \Concept{non-smiling} to \Concept{smiling} \cite{esser2020disentangling}; or a truly linear change of an
%
object's representation 
in latent space when continuously modifying 
rotation or color. 
    %
%
And lastly, it is not taken for granted that local approximation by a trajectory with 1D curvature (i.e., a straight line) is strong enough to capture all concept information of interest. Several approaches now discard this assumption by using the highly non-straight trajectories of traversing concepts in generative model latent spaces \cite{esser2020disentangling,varshney2024generating}. 
Validating and addressing the linearity assumption is essential for developing more flexible, potentially non-linear concept representations that accurately capture the complexities of real-world data.

\subsubsection{Richer Ontological Commitment of Concept Representations.}


So far, C-XAI research has mostly focused on the ability of a model to grasp a concept as intended. Unfortunately, little work beyond this is devoted to \Challenge{systematic investigation of the ontological commitment in trained models, in particular the relations that they can model}. In the context of knowledge embeddings, on the other hand, there exist principled embedding approaches that capture rich concept relations, including subsumption \cite{gutierrez-basulto18from,xiong_faithful_2022,mena-ki:22}, or negation \cite{oezcep-alc-JAIR:23}.

In consideration of models that exhibit reasoning capabilities, further requirements may arise. This has already been shown for the usual approach of geometric containment in latent space to represent concept subsumption, i.e., points inside a region represent the instances of some concept. 
In order to allow reasoning, concept regions cannot be arbitrarily shaped \cite{leemhuis-oezcep-JAR:23}. Put differently, the geometry of concepts and their relations in latent space is tightly coupled with the reasoning capabilities that can be achieved. Little of geometry-reasoning interdependency has been revealed so far. A promising direction to solve this issue is to \Proposal{investigate whether vision models use some of the known principled embedding approaches} such as from spatial reasoning \cite{vanharmelen2007handbook,dylla2017survey,oezcep-alc-JAIR:23} 
and how to extend existing C-XAI approaches to extract these.


Another open challenge is to \Challenge{develop tools for verification of a model's ontological commitments}. Options for achieving this are to (a) find accurate representations of known relations in the model, or (b) verify a given relation function commitment (like the cosine similarity) against expected behavior. 
An approach to the first challenge could be \Proposal{considering so-called reification of relations} \cite{mena-ki:22}, an idea from knowledge graph embeddings that flexibly represents relations themselves as concepts.
For the second challenge, \Proposal{both the rich common sense ontologies should be taken into account, complemented by densely labeled visual datasets labeling both concepts and local relations}, similar to scene graph datasets \cite{krishna2017visual}. Note that this might require considerable efforts in the community to \Proposal{define task-specific sub-ontologies, and to develop more specialized and controllable datasets and testing environments}, such as 3D-generated scenes with automatic annotations~\cite{infinigen2023infinite,infinigen2024indoors} or generative AI-produced data. 
Apart from that, understanding what relations a deep neural network of given depth can accurately model, and investigating whether it does model the relations of interest, are an important step for fully understanding the automatic reasoning applied by the model.

\section{Concept Control}
\label{sec:concept-control}

Concept-based explanations not only allow us to understand a model, but also provide us with means to change the model to achieve a specific objective, e.g., improving the generalizability of the model. Among several ways of changing a model (e.g., improving the quality of the training set, choosing a better model architecture, or directly modifying the intermediate representations of an input), we see high potential in targeted modification of the intermediate representations of model inputs, which we refer to as \emph{concept control} or \emph{concept intervention}.

\subsection{Modification of Latent Representations}


In the C-XAI literature, concepts are often represented as vectors in an embedding space \cite{kim2018interpretability,oikarinen_label-free_2023,abid_meaningfully_2022,yuksekgonul_post-hoc_2023}. Given a set of concepts $\mathcal{C} = \{c_{1}, \dotsc, c_{n}\}$, we can regard the corresponding concept vectors $w_{c_{1}}, \dotsc, w_{c_{n}} \in \mathbb{R}^{d}$ as a 
generating set or, by abusing the language, a basis of interpretable linear subspace of the embedding space $\mathbb{R}^{d}$. The $i$\textsuperscript{th} coordinate of a representation with respect to that \emph{concept basis
} captures the strength of the presence of concept $c_{i}$ in the representation. 
This allows us to intervene on a concept $c_{i}$ in an intuitive way: increasing (or decreasing) the $i$\textsuperscript{th} coordinate leads to increasing (or decreasing) the presence of concept $c_{i}$ in the representation.
Koh et al.~\cite{koh2020concept} were to our knowledge the first to apply a concept intervention. This was done on the CBM architecture, where a complete layer is trained such that each single neuron, which corresponds to a unit vector in the embedding space, is associated with a given concept in an ante-hoc supervised manner.
An issue here is the need for ground-truth concept labels, which are often unavailable.

Several approaches circumvent the above issue by interpreting the concept vectors in a post-hoc manner. Abid et~al.\ \cite{abid_meaningfully_2022} and Yuksekgonul et~al.\ \cite{yuksekgonul_post-hoc_2023} use the CAVs of a pre-trained vision model as a concept basis and identify concepts that need to be added (or strengthened) or removed (or suppressed). To apply local interventions, Abid et al.\ learn counterfactual explanations, that is, identifying the concepts that should have been added or removed so that the model predicts the correct label. In contrast to that, the post-hoc CBM approach~\cite{yuksekgonul_post-hoc_2023} intervenes globally by removing a spurious concept to predict a class, e.g., removing the concept \Concept{dog} to predict the concept \Concept{table} in the test set when \Concept{dog} is spuriously correlated with \Concept{table} in the training set, but not in the test set. Further methods include the editing of classifiers \cite{santurkar_editing_2021} that can control the behavior of an image classifier by using only a single example, P-ClArC \cite{anders_finding_2022}, which projects out concept directions, and RR-ClArC \cite{dreyer_hope_2024}, which regularizes CAVs during training to guide the model to become less reliant on biases.
Similarly to CBMs, post-hoc C-XAI methods require a predefined set $\mathcal{C}$ of concepts.

\subsection*{Open Challenges}
\label{sec:cc-open-challenges}

We identify three underexplored areas: imposing logic constraint, application of concept control, and mitigation of side effects which are explained below.

\subsubsection{Imposing Logical Constraints.} Logical constraints can be imposed on concepts and allow for tight neurosymbolic integration~\cite{lee_neural_2024,donadello2017logic,schwalbe2022enabling}. Such logical constraints can be used to guarantee the consistency of the model's reasoning \cite{schwalbe2022enabling} and to align the model with a knowledge base or with different criteria by the user \cite{donadello2017logic} (e.g., criteria related to ethics, privacy, and safety). C-XAI brings in the benefit that one can directly act on concepts in the embedding space of intermediate layers of a pretrained model instead of on the model's output (e.g., removing \Concept{skin color}, which is usually not an output label, from an intermediate layer for fairness assessment). Controlling intermediate representations promises to achieve higher coverage of concepts and performance and greater flexibility.

We highlight two future challenges on how such constraints can be enforced: (i)~by \Challenge{guiding the model training} and (ii) \Challenge{globally modifying intermediate representations}. For the first challenge, one can use a \Proposal{multi-task training routine that simultaneously or alternatingly updates the concept models} to maintain a correct association of the concepts\gsadd{,} and updates the model parameters according to both the main task and the constraints on the concepts. 
Constraints may be formulated and approximated using regularization terms, as proposed in the semantic loss formulation in \cite{xu_semantic_2018,badreddine_logic_2022}. In contrast to this \enquote{soft} approach to model weights for the first challenge, the second challenge can be tackled by inserting intermediate processing steps that will modify the intermediate input representations to comply with the constraints, for example, by \Proposal{linear projection or linear skew.} A proof of concept is shown in \cite{ribeiro_modifying_2023}, but this was on simplistic unit-vector concepts. Generally, \Challenge{it remains to be shown that logical constraints can be applied to diverse image datasets and with varying expressivity of the logical constraints}, for example, allowing relations or functions in addition to concepts in the logical constraints. 
    
Note that the logical constraints imposed on the model need to be compatible with the rules that can be extracted from the same model (cf.~\autoref{sec:conc-repr}). More precisely, depending on the expressivity \cite{schwalbe2022enabling} of the logic language used to extract the rules from the model, logical constraints of high or lower expressive power can be imposed on the model. Therefore, investigating the reasoning of the model can affect concept control.

\subsubsection{Applications of Concept Control.}

The motivation for concept control can come from various applications, such as model editing and debugging, increasing the robustness of models against adversarial attacks or distribution shifts~\cite{liang_metashift_2022}, or \Challenge{avoiding catastrophic forgetting (i.e., retaining previously learned knowledge) in new tasks in a lifelong learning scenario}~\cite{wang2024comprehensive}. For example, a self-driving car that was trained to recognize humans based on specific clothes may fail in areas with a different climate or culture. Although model editing and debugging are performed on a model after training, catastrophic forgetting can already be mitigated at training time by regularizing by \Proposal{penalizing deviation of the model's ontological commitment across different tasks}.

Evaluating C-XAI methods is still an open problem~\cite{poeta_concept-based_2023}. Approaching C-XAI from the perspective of concept control with \Proposal{concrete objectives in applications, such as model correction~\cite{dreyer_hope_2024}, can be an effective way to evaluate C-XAI methods}. Therefore, \Challenge{identifying what potential applications can benefit from concept control and comprehensively evaluating the controllability of C-XAI methods in such applications} can be beneficial for the C-XAI research.

\subsubsection{Mitigating Side Effects.}
\label{sec:mitig-side-effects}

Concepts can depend on each other; for example, in most cases the concept \Concept{car} co-occurs in an image with \Concept{wheel} which again includes the concept \Concept{round}. Thus, modifying a concept globally can affect other concepts, for example, replacing \Concept{round} with \Concept{rectangular} affects concepts \Concept{wheel} and \Concept{car}. This side effect, which is also known as the ripple effect in the language model editing literature~\cite{cohen_evaluating_2024}, can be a big impediment to controlling concepts. \Challenge{Identifying the side effects of a specific concept control mechanism and avoiding the side effects} are therefore important open challenges. \Proposal{Inspirations from C-XAI approaches to natural language processing} (e.g., \cite{cohen_evaluating_2024}) could be a starting point for next steps.

\section{Conclusion}
    In this paper, we have examined the current state and open challenges in C-XAI for CV, focusing on concept types, expressive representations, and use of control.
     We identified three currently underexplored areas with high potential to advance the field:
    \begin{compactenum}[(i)]
        \item \highlight{Expand the types of concepts} that can be extracted and analyzed to temporal ones, as well as recent model architectures like ViTs.
        \item Inspired by knowledge representation, develop \highlight{richer concept representations} that go beyond simple point-estimate vector embeddings and capture the complexity and relations of concepts learned by CV models.
        \item On the application side, improve the techniques for concept control by \highlight{imposing logical constraints} directly on the model's internal representations.
    \end{compactenum}
    Addressing these challenges, C-XAI methods can provide deeper insights into the inner workings of vision models and enable more fine-grained, interactive control over their behavior. This will be crucial for the verification and maintainability of critical CV applications.
    %
    %
    We hope to have provided a good starting point for researchers new to the field, as well as helpful inspiration for the community to advance it further.


\section*{Acknowledgements}
Jae Hee Lee and Stefan Wermter gratefully acknowledge support from the German Research Foundation DFG for the project CML TRR169.

\bibliographystyle{splncs04} %
\bibliography{main}

\clearpage
\setcounter{section}{0}
\renewcommand{\thesection}{\Alph{section}}

\section{Appendix}

\subsection{Taxonomy of C-XAI Methods}\label{sec:appendix-taxonomy}

\begin{figure}[h]
    \centering
    {

\begingroup%
\colorlet{leaf}{magenta}%
\colorlet{supervised}{ForestGreen!40}%
\colorlet{unsupervised}{CornflowerBlue!50}%
\centering%
\tiny%
\tikzset{
    my node/.style={
      draw=gray,
      inner color=gray!5,
      thin,
      rounded corners=3,
      text height=1.3ex,
      text depth=0ex,
      font=\sffamily,
      drop shadow,
      inner sep=2pt,
      text centered,
    },
    root node/.style={
        my node,
        draw=none,
        font=\itshape{\color{gray}$\bullet$},
        rotate=90,
    },
    leaf node/.style={
      my node,
      draw=leaf,
      font=\sffamily\color{black},
      drop shadow={fill=leaf!10}
    },
    supervised/.style={inner color = supervised, outer color = supervised},
    supervised posthoc/.style={inner color = supervised!80, outer color = supervised!80},
    supervised finetuning/.style={inner color = supervised!50, outer color = supervised!50},
    supervised antehoc/.style={inner color = supervised!20, outer color = supervised!20},
    unsupervised/.style={inner color = unsupervised!50, outer color = unsupervised!50},
    unsupervised antehoc/.style={inner color = unsupervised!20, outer color = unsupervised!20},                                                      unsupervised posthoc/.style={inner color = unsupervised!80, outer color = unsupervised!80},
  }%
\forestset{
     default preamble={
     for tree={
        my node,
        grow'=east, align=left, if n children=0{tier=last}{},
        l sep+=5pt, s sep=1.5pt,
        parent anchor=east,
        child anchor=west,
        edge={gray},
        edge path={\noexpand\path [draw, \forestoption{edge}, ] (!u.parent anchor) -- +(10pt,0) |- (.child anchor)\forestoption{edge label};},
     }
 }
}%
\begin{NoHyper}
\begin{forest}
    [{\textbf{C-XAI methods} \cite{schwalbe_concept_2022,poeta_concept-based_2023,lee_neural_2024,rauker2023transparent}}, 
    root node, font=\normalfont, parent anchor=south,child anchor=north,align=left,
      [supervised, supervised
        [post-hoc, supervised posthoc
          [global, supervised posthoc
              [single unit, supervised posthoc
                [{NetDissect \cite{bau2017network}}, leaf node, supervised posthoc]
                [{GANDissect \cite{bau2018gan}}, leaf node, supervised posthoc]
              ]
              [linear model, supervised posthoc
                [classification, supervised posthoc
                  [{TCAV \cite{kim2018interpretability}}, leaf node, supervised posthoc]
                ]
                [segmentation, supervised posthoc
                  [{Net2Vec \cite{fong2018net2vec}, \cite{schwalbe2021verification}}, leaf node, supervised posthoc]
                  [{CLM \cite{lucieri2020explaining} (TCAV+feat.\,imp.)}, leaf node, supervised posthoc]
                ]
                [regression, supervised posthoc
                  [RCV \cite{graziani2018regression,graziani2020concept}, leaf node, supervised posthoc]
                ]
              ]
              [kernel-based model, supervised posthoc
                [{TCAR \cite{crabbe2022concept}}, leaf node, supervised posthoc]
              ]
          ]
          [local, supervised posthoc
              [{GCPV \cite{mikriukov2023gcpv}}, leaf node, supervised posthoc]
              [{Derivatives \cite{zhang2018examining}}, leaf node, supervised posthoc]
          ]            
        ]
        [via finetuning, supervised finetuning
          [{Concept Whitening \cite{chen2020concept}}, leaf node, supervised finetuning]
        ]
        [ante-hoc, supervised antehoc
          [fully supervised, supervised antehoc
              [vanilla bottleneck, supervised antehoc
                [classification, supervised antehoc
                  [{CBM \cite{koh2020concept}}, leaf node, supervised antehoc]
                ]
                [segmentation, supervised antehoc
                  [{Semantic Bottlen.\ \cite{losch2019interpretability}}, leaf node, supervised antehoc]
                ]
              ]
              [{Denoised\,\cite{bahadori2020debiasing} and AR\,\cite{havasi2024addressing} CBM}, leaf node, supervised antehoc]
              [{CEM \cite{espinosazarlenga2022concept}}, leaf node, supervised antehoc]
          ]
          [semi-supervised, supervised antehoc
            [{Weakly Supervised CBM \cite{belem2021weakly}}, leaf node, supervised antehoc]
            [{Label-free CBM \cite{oikarinen_label-free_2023} (CLIP labels)}, leaf node, supervised antehoc]
          ]
          [partly supervised, supervised antehoc
            [{CBM-AUC \cite{sawada2022concept}}, leaf node, supervised antehoc]
          ]
        ]
      ]
      [unsupervised, unsupervised posthoc
        [post-hoc, unsupervised posthoc
          [single-layer features, unsupervised posthoc
            [standard clustering, unsupervised posthoc
              [{ACE \cite{ghorbani2019towards,yeh2020completenessaware}}, leaf node, unsupervised posthoc]
              [{VRX \cite{ge2021peek}}, leaf node, unsupervised posthoc]
            ]
            [matrix factorization, unsupervised posthoc
              [{ICE \cite{zhang2021invertible}}, leaf node, unsupervised posthoc]
              [{CRAFT \cite{fel2023craft,fel_holistic_2023} (ACE+ICE)}, leaf node, unsupervised posthoc]
              [{DMA\&IMA \cite{leemann2023when}}, leaf node, unsupervised posthoc]
            ]
            [subspace clustering, unsupervised posthoc
              [{MCD \cite{vielhaben2023multidimensional}}, leaf node, unsupervised posthoc]
            ]
          ]
          [multi-layer features, unsupervised posthoc
            [{ECLAD \cite{posada-moreno2024eclad,posada-moreno2023scalable}}, leaf node, unsupervised posthoc]
          ]
        ]
        [ante-hoc, unsupervised antehoc
          [classification, unsupervised antehoc
            [per-class concepts, unsupervised antehoc
              [{ProtoPNet \cite{chen2019this,li2018deep}}, leaf node, unsupervised antehoc]
              [{ProtoPNeXt \cite{willard2024this}}, leaf node, unsupervised antehoc]
            ]
            [shared concepts, unsupervised antehoc
              [{ProtoPShare \cite{rymarczyk2021protopshare}}, leaf node, unsupervised antehoc]
            ]
          ]
          [object detection, unsupervised antehoc
            [{CSPP \cite{feifel2021reevaluating}}, leaf node, unsupervised antehoc]
          ]
        ]
      ]
    ]
\end{forest}%
\end{NoHyper}
\endgroup%
}
    \caption{Detailed taxonomy of state-of-the-art C-XAI methods.}
    \label{fig:taxonomy}
\end{figure}
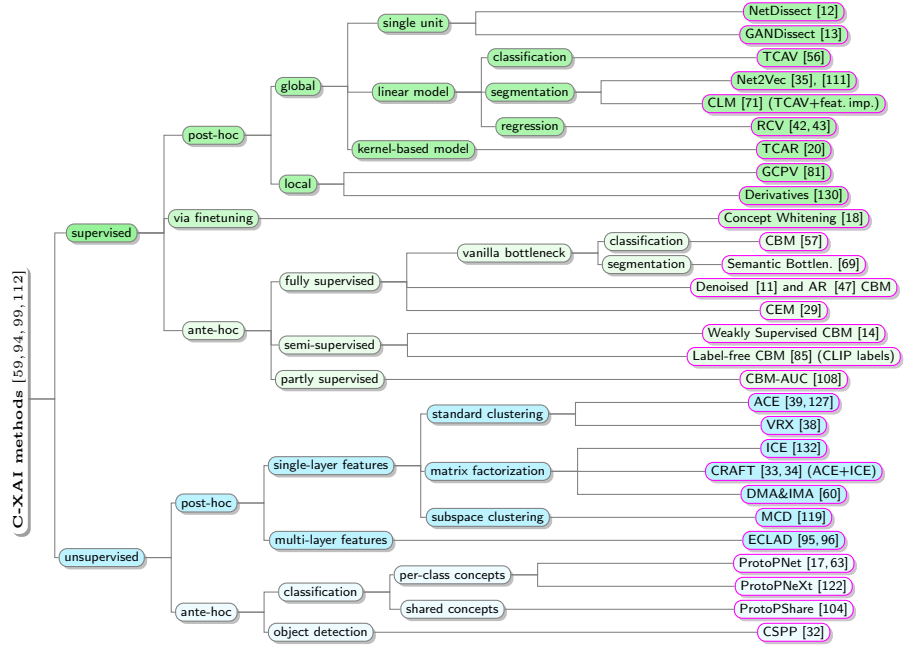

\subsection{Details on the Illustration of Concept Distribution}\label{sec:appendix-graphics}

\begin{figure}
    \centering
    \includegraphics[width=\linewidth]{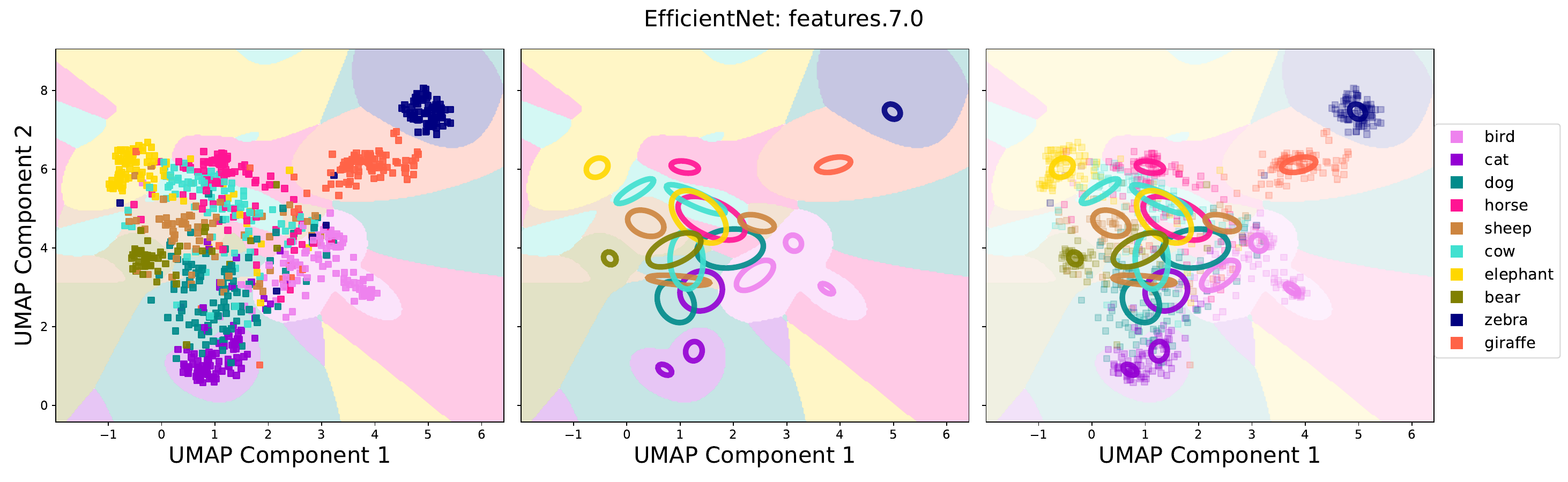}
    \caption{Some creation steps of \autoref{fig:gcpv-distribution}, from \emph{left} to \emph{right}: (1) Given the local concept embedding vectors, apply UMAP for dimensionality reduction; (2) fit Gaussian mixture models and determine boundaries of standard deviations; (3) add background shading to indicate most probable concepts (here shown for all 3 graphics), and overlay everything.}
    \label{fig:gcpv-distribution-detailed}
\end{figure}

The illustration of the concept distribution in EfficientNet-B0 from \autoref{fig:gcpv-distribution}, shown in its separate steps in \autoref{fig:gcpv-distribution-detailed}, was obtained as follows:

\begin{enumerate}
    \item
    The local concept vectors are obtained using the \highlight{GCPV} optimization technique suggested in \cite{mikriukov2023gcpv} with the difference that the optimization objective was changed to pseudo-BCE-loss like in~\cite{fong2018net2vec}:
    Given an image together with a concept label, a linear classifier is optimized to correctly classify the activation map pixels of this single image as a concept or background. The normal vector of this linear mapping is then taken as the concept embedding vector for this concept local to this image.
    This is essentially a local version of Net2Vec~\cite{fong2018net2vec}, where the resulting vectors represent concepts in the context (background) for each sample.
    \item 
    This procedure was applied to the concepts shown belonging to the supercategory \enquote{animal} of the MS~COCO dataset~\cite{lin2014microsoft}, and the activations of the last layer of \texttt{ features.7.0} of \highlight{EfficientNet-B0}~\cite{tan2019efficientnet}.
    \item
    The reduced dimensionality points shown are the density-preserving $L_2$-distance \highlight{UMAP}~\cite{mcinnes2020umap} 2d-mapping of the set of embedding vectors of the local image concept (of all concepts).
    \item
    On each concept's vectors separately, a \highlight{multivariate Gaussian mixture model} was trained to capture a 2d representation of their distribution.
    The number of components was determined using the Bayesian information criterion (BIC), using 1 to 3 components and ignoring outliers.
    
    Note that despite doing the (non-distance preserving) UMAP mapping first and only after that the distribution approximation, the diagrams can still be considered informative, as UMAP preserves the local density information, and thus separation of modes.
    \item
    \highlight{Ellipses} are used to visualize the per-concept fitted multivariate Gaussian distributions:
    They demarcate the boundary at one standard deviation of each of the Gaussian components.
    \item
    The \highlight{background color} marks are the most probable concept at that point according to the fitted Gaussians.
\end{enumerate}



\end{document}